\documentclass{article}

\usepackage{arxiv}
\usepackage[utf8]{inputenc} 
\usepackage[T1]{fontenc}    
\usepackage{hyperref}       
\usepackage{url}            
\usepackage{booktabs}       
\usepackage{amsfonts}       
\usepackage{nicefrac}       
\usepackage{microtype}      

\usepackage{amsmath}
\usepackage{amssymb}
\usepackage{mathtools}
\usepackage{amsthm}
\usepackage{amsfonts}
\usepackage{bm}
\usepackage{float}
\usepackage{geometry}
\usepackage[capitalize,noabbrev]{cleveref}
\usepackage{multirow}
\usepackage[misc]{ifsym}
\usepackage{graphicx}
\usepackage{subcaption}

\theoremstyle{plain}

\theoremstyle{definition}

\theoremstyle{remark}

\usepackage{algorithm}
\usepackage{algorithmic}

\def \MethodName {Marope\xspace}
\usepackage{xspace}
\usepackage{wrapfig}

\title{Cooperative Long Rope Skipping via Multi-Agent Reinforcement Learning}

\author{
Zihao Wang$^{1,2}$, Shijie Peng$^{2}$, Kerui Wu$^{2}$, Yu Huang$^{2}$, Ruiqi Xue$^{1,2}$,\\
\textbf{Dong Liu$^{3}$, Tian Xu$^{1,2}$, Lei Yuan$^{1,2}$, Yang Yu$^{1,2}$}\\
$^1$ National Key Laboratory of Novel Software Technology, Nanjing University, Nanjing, China\\
$^2$ School of Artificial Intelligence, Nanjing University, Nanjing, China\\
$^3$ Beijing Academy of Artificial Intelligence, BAAI, Beijing, China\\
\texttt{\{wangzh,xuerq,xut,yuanl,yuy\}@lamda.nju.edu.cn},\\
\texttt{\{pengsj,wukerui,huangy\}@smail.nju.edu.cn}, \texttt{liudong@baai.ac.cn}
}

\begin{document}
\date{}
\maketitle
\begin{abstract}
    Humans exhibit remarkable motor agility, enabling a wide range of dynamic skills such as running and jumping, which highlights the great potential of humanoid robots for athletic locomotion. Among athletic sports, long rope skipping requires two rope turners to cooperatively swing the rope while adapting to a player under different jumping rhythms, making it a meaningful yet challenging task for humanoid robots. Although existing methods for humanoid sports have achieved success in single-agent and interaction-free settings, such as running, dancing, and parkour, task scenarios that require precise coordination among multiple participants remain largely unexplored.
    To this end, we propose \MethodName, a multi-agent reinforcement learning (MARL) framework for cooperative long rope skipping with multiple humanoid robots. Specifically, \MethodName adopts a hierarchical reinforcement learning framework for policy training. At the lower level, it learns decentralized rope manipulation policies through MARL, while at the upper level, a centralized scheduling policy is trained to coordinate the execution of the lower-level policies. To improve generalization across different player behavioral styles, \MethodName further incorporates diverse jumping policies into cooperative game training.
    We evaluate our approach on Unitree G1 humanoid robots in both simulation and real-world settings. Experimental results demonstrate that \MethodName outperforms various baselines, achieving more efficient and stable rope manipulation as well as more robust and adaptable cooperation with varied players. More results can be found on the project website: \href{https://marope-dev.github.io/}{\texttt{https://marope-dev.github.io/}}.
\end{abstract}


\section{Introduction}
Developing humanoid robots capable of emulating agile and robust human motions across diverse and complex environments has long been a central goal of embodied intelligence~\cite{hirai1998development, tong2024advancements, he2024learning}. Within this broader pursuit, research on humanoid sports~\cite{crowley2023optimizing,qi2023vertical,qin2018music,zhuang2025humanoid,su2025hitter} has emerged as a particularly active direction, driven by the widespread popularity and unique recreational value of sports in human life, as well as the strong demands they impose on versatile and dynamic whole-body control. Among these sports, long rope skipping is a widely practiced recreational activity, valued for its highly social nature which requires multiple participants and its benefits for physical fitness. Enabling humanoids to participate in long rope skipping therefore presents an important yet challenging research problem.

Although humanoid sports have received increasing attention, most existing studies focus on single-agent task settings, where a centralized learning framework is employed to control an individual humanoid for an isolated athletic skill. However, long rope skipping departs from this formulation in several fundamental aspects. First, the task requires two humanoid rope turners to generate a coherent rope motion through whole-body control, while maintaining balance and stable foot contacts. Second, the rope is a deformable and underactuated object whose intermediate configuration cannot be directly controlled~\cite{liu2023robotic}, making simple trajectory replay~\cite{bruce2017one,di2024effectiveness} or blind motion tracking~\cite{thuruthel2018stable,liao2025beyondmimic} insufficient for reliable rope manipulation. Third, successful skipping depends on precise temporal synchronization between the rope rotation phase and the player’s jumping phase; even small phase mismatches may result in rope-player collisions or failed jumps. In open-world cooperative scenarios, this challenge becomes even more pronounced, as players may exhibit diverse and unknown behavioral styles~\cite{yuan2023survey}. Therefore, cooperative long rope skipping calls for a learning framework that can coordinate multiple humanoids, manipulate a flexible rope in a closed loop and adapt to diverse jumping behaviors.

To address these challenges, we propose \MethodName, a multi-humanoid cooperative reinforcement learning (RL) framework for long rope skipping that can adapt to diverse rope jumping patterns. Specifically, \MethodName first pretrains a decentralized rope manipulation policy for two humanoid rope turners with MAPPO under the CTDE paradigm, using a compact command space that specifies the rotation center and rotation angular velocity instead of prescribing full rope trajectories. Built on this reusable low-level skill, \MethodName learns a centralized scheduling policy to dynamically adjust the manipulation commands to synchronize with a player's jumping rhythm while reducing rope-player collisions. To improve robustness to different partners, \MethodName further trains a latent-conditioned jumping policy with an IPM-based diversity intrinsic objective and uses the resulting diverse behaviors for data augmentation, encouraging the scheduling policy to generalize across varied jumping styles. We evaluate our approach on Unitree G1 humanoid robots in both simulation and real-world environments. Experimental results demonstrate that our method outperforms various baselines, and produces more efficient and stable rope manipulation as well as more robust and adaptable cooperation with diverse players. To the best of our knowledge, this work presents the first multi-humanoid cooperative long rope skipping system, extending humanoid sports from single-agent athletic skills to tightly coordinated multi-robot scenarios.

\section{Related Work}
\paragraph{Learning-based Humanoid Control}
Learning-based methods have substantially advanced humanoid control, particularly in robust locomotion and sim-to-real transfer. Early works focused on stable and versatile bipedal locomotion, including walking, running, velocity tracking, and terrain adaptation \cite{li2019using, li2021reinforcement, radosavovic2024real, radosavovic2024humanoid, long2025learning}. Recent benchmarks and systems further extend humanoid control toward high-dimensional whole-body tasks, highlighting both the promise and challenges of learning general-purpose humanoid behaviors \cite{sferrazza2024humanoidbench}. Meanwhile, motion imitation and teleoperation have enabled humanoid robots to acquire expressive whole-body skills from human motion data or demonstrations, such as dancing, gesturing, dexterous teleoperation, and dynamic motion tracking \cite{liao2025beyondmimic,cheng2024expressive, ji2025exbody2, he2025omnih2o}. Building on these advances, humanoid sports have emerged as a compelling testbed for agile whole-body control. Prior work has explored vertical jumping, parkour, table tennis, badminton, and tennis, demonstrating increasingly dynamic interactions with environments, objects, and human players \cite{qi2023vertical, zhuang2025humanoid, su2025hitter, chen2026learning, zhang2026learning}. However, most methods are formulated as single-humanoid control problems, where one policy controls an individual robot to perform an athletic skill or react to external objects. In contrast, long rope skipping requires multiple humanoids to coordinate whole-body motions through a shared deformable rope while synchronizing with a jumping participant.

\paragraph{Multi-agent Reinforcement Learning}
Multi-agent Reinforcement Learning (MARL) utilizes RL to address multi-agent problems. Unlike single-agent settings, MARL faces the curse of dimensionality in the joint state-action space, which stems from the growing number of agents. To overcome this challenge, typical works utilize value decomposition~\cite{sunehag2017value-vdn,rashid2018qmix} or decentralized policy gradient~\cite{lowe2017multi,wang2021dop,yu2022surprising} to transform the complex high-dimensional joint space into tractable low-dimensional representations, demonstrating high learning efficiency in fields such as autonomous driving~\cite{zhang2024multi}, financial trading~\cite{fang2023learning}, and embodied intelligence~\cite{feng2026multi}. In addition to the aforementioned methods and their variants, many other research directions have been explored in MARL, including efficient communication mechanisms for mitigating partial observability under decentralized policy execution~\cite{zhu2022survey}, offline policy learning~\cite{zhang2023discovering}, world models for MARL~\cite{wang2022model}, and policy robustness in the presence of perturbations~\cite{guo2022towards}.

\section{Preliminaries}

We formalize the cooperative long rope skipping problem as a Partially Observable Markov Game $\mathcal{M} = \langle \mathcal{N}, \mathcal{S}, \{ \mathcal{A}^{i} \}_{i \in \mathcal{N}}, P, \{ R^{i} \}_{i \in \mathcal{N}}, \gamma, \{ \Omega^{i} \}_{i \in \mathcal{N}}, \mathcal{O} \rangle$, where $\mathcal{N} = \{ 1, 2, \cdots, n \}$ represents the set of participant agents, $\mathcal{S}$ is the global state space, $\mathcal{A}^{i}$ and $\Omega^{i}$ are the action and observation spaces of agent $i$. At timestep $t$ with global state $s_{t} \in \mathcal{S}$, each agent $i$ observes a partial observation $o_{t}^{i} \in \Omega^{i}$ according to the observation probability $O(\cdot \mid s_{t}, i)$ and chooses an action $a_{t}^{i}$. After executing the joint action $\boldsymbol{a}_{t} = \{ a_{t}^{i} \}_{i \in \mathcal{N}}$, each agent $i$ will receive a reward $r_{t}^{i} = R^{i}(s_{t}, \boldsymbol{a}_{t})$ and the global state will transit to $s_{t + 1} \sim P(s_{t + 1} \mid s_{t}, \boldsymbol{a}_{t})$. The goal of each agent $i$ is to learn a policy $\pi^{i}(a_{t}^{i} \mid o_{t}^{i})$ that maximizes its expected return $\mathbb{E} \left[ \sum_{t = 0}^{\infty} \gamma^{t} r_{t}^{i} \right]$ under discounted factor $\gamma \in [0, 1)$.


\section{Method}

\begin{figure}
	\centering
	\includegraphics[width=\linewidth]{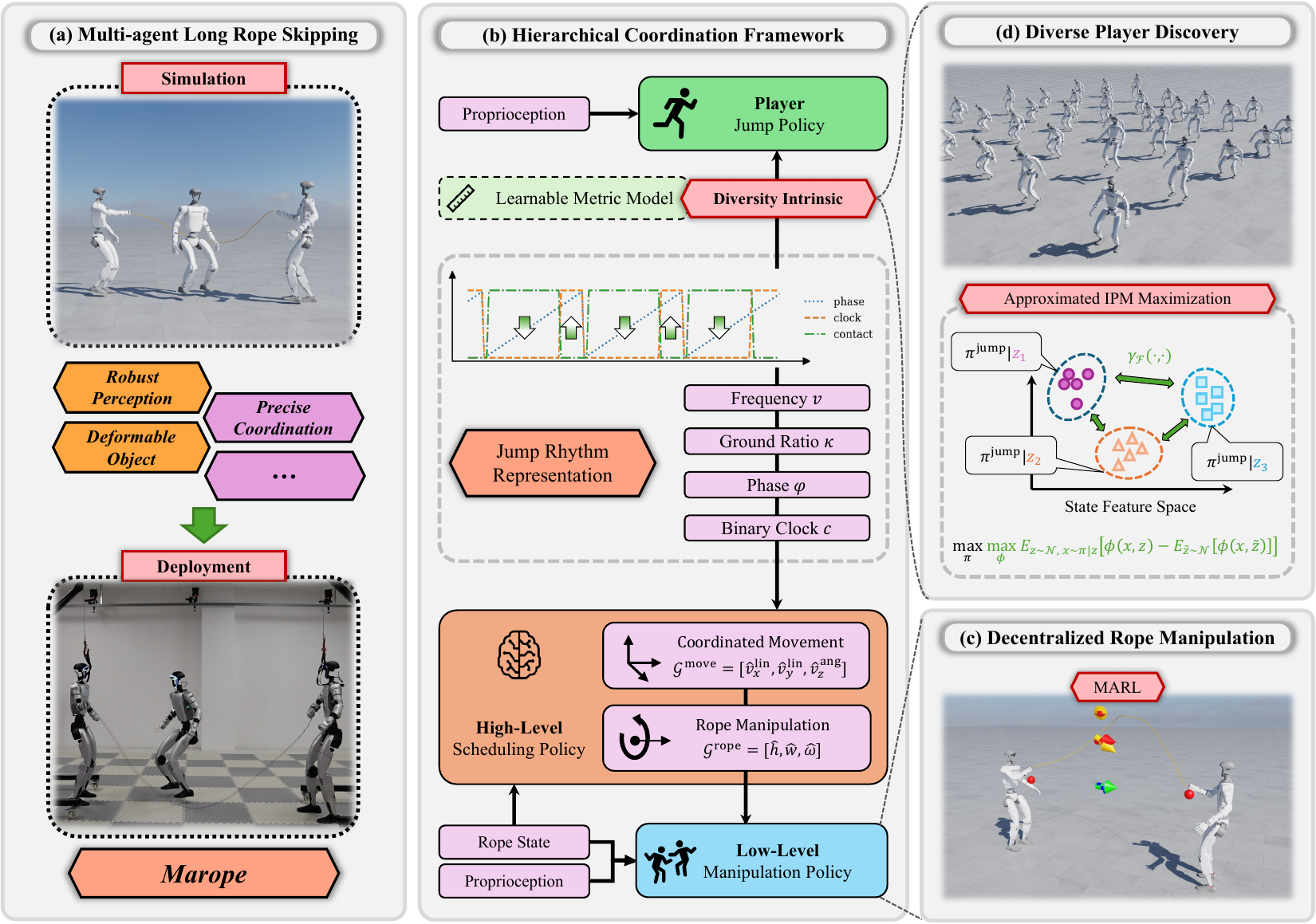}
	\caption{\textbf{Overview of \MethodName}. (a) For long rope skipping task, \MethodName builds a pipeline for learning long rope skipping skills on multiple humanoid robots (b) A hierarchical coordination framework is used for efficient coordination with player under specific jump rhythm. (c) The low-level decentralized rope manipulation policy is trained via MARL. (d) Through an IPM-based diversity intrinsic objective, diverse player behaviors are discovered to improve generality of high-level scheduling policy.}
	\label{fig:main}
\end{figure}

This section gives the detailed \MethodName, a novel framework for learning cooperative long rope skipping. \cref{sec:low_level} presents the formulation and decentralized training of the low-level rope manipulation policy, \cref{sec:high_level} introduces rhythm representation and training of the high-level centralized scheduling policy, while \cref{sec:diverse_jump} describes how \MethodName discovers diverse player behavior with a diversity intrinsic objective to improve policy generalization and adaptability.

\subsection{Decentralized Cooperative Rope Manipulation}
\label{sec:low_level}

Rope manipulation is the foundation of the long rope skipping, thus, we first pretrain a decentralized cooperative rope manipulation policy $\pi^{\text{manip}}(a_{t} \mid o_{t}^{\text{manip}})$ for two humanoid rope turners. The core objective is to rotate the rope around a reference center $\mathbf{p}^{\text{c}}$ under a target angular velocity $\hat{\boldsymbol{\omega}}$.

\paragraph{Rope Manipulation Modeling}
Specifically, let $\mathbf{p}_{\text{base}}^{1}$ and $\mathbf{p}_{\text{base}}^{2}$ denote the base link positions of two humanoid rope turners, we define the reference center $\mathbf{p}^{\text{c}}$ and rotation axis $\mathbf{e}^{\text{r}}$ as follows:
\begin{equation}
    \mathbf{p}^{\text{c}} = \Pi_{xy} \left( \frac{\mathbf{p}_{\text{base}}^{1} + \mathbf{p}_{\text{base}}^{2}}{2} \right) + [0, 0, \hat{h}]^{\top}, \quad \mathbf{e}^{\text{r}} = \frac{\Pi_{xy}(\mathbf{p}_{\text{base}}^{2} - \mathbf{p}_{\text{base}}^{1})}{\lVert \Pi_{xy}(\mathbf{p}_{\text{base}}^{2} - \mathbf{p}_{\text{base}}^{1}) \rVert_{2}},
\end{equation}
where the horizontal projection operator $\Pi_{xy}([x, y, z]^{\top}) = [x, y, 0]^{\top}$, $\hat{h}$ is the target rotation height. The target rotation angular velocity can be further given as $\hat{\boldsymbol{\omega}} = \hat{\omega} \cdot \mathbf{e}^{\text{r}}$ via a signed scalar $\hat{\omega}$. This definition guarantees $\text{SE}(2)$ mobility for the rope manipulation behavior, enabling the humanoid rope turners to swing the rope while simultaneously translating, turning, and repositioning.

Accordingly, we design the command as $\mathcal{G}^{\text{manip}} = \left[ \mathcal{G}^{\text{move}}, \mathcal{G}^{\text{rope}} \right]$, including coordinated movement related components $\mathcal{G}^{\text{move}} = \left[ \hat{v}_{x}^{\text{lin}}, \hat{v}_{y}^{\text{lin}}, \hat{v}_{z}^{\text{ang}} \right]$ and rope manipulation related components $\mathcal{G}^{\text{rope}} = \left[ \hat{h}, \hat{w}, \hat{\omega} \right]$, where $\hat{v}_{x}^{\text{lin}}, \hat{v}_{y}^{\text{lin}}, \hat{v}_{z}^{\text{ang}}$ refer to the target linear and angular velocity of the reference center and rotation axis in the world frame, $\hat{w}$ is the target width for two rope ends. Compared with explicitly specifying reference trajectories for a deformable object, the above abstraction provides a compact description for the key rope motion pattern in the long rope skipping task: $\hat{v}_{x}^{\text{lin}}, \hat{v}_{y}^{\text{lin}}, \hat{v}_{z}^{\text{ang}}$ control the overall stances, $\hat{h}$ and $\hat{w}$ constrain the region swept by rope, whereas $\hat{\omega}$ specifies the swinging rhythm.

\paragraph{Symmetric Observation Construction}
Given the modeling above, the observation of the rope manipulation policy on humanoid rope turner $i \in \{ 1, 2 \}$ comprises three parts: command observation $o_{t}^{\text{cmd}, i}$, rope morphology history $o_{t - H + 1 : t}^{\text{rope}, i}$ and proprioceptive sensing history $o_{t - H + 1 : t}^{\text{proprio}, i}$. For simplicity, we omit the subscript $t$ in the following detailed explanation.
The command observation is computed as $o^{\text{cmd}, i} = \left[ \hat{v}_{x}^{\text{lin}, i}, \hat{v}_{y}^{\text{lin}, i}, \hat{v}_{z}^{\text{ang}}, \hat{h}, \hat{w}, \operatorname{sign}(i) \cdot \hat{\omega} \right]$, where the target linear velocity of the reference center $\left( \hat{v}_{x}^{\text{lin}, i}, \hat{v}_{y}^{\text{lin}, i} \right)$ is represented in the yaw-only base link frame of humanoid rope turner $i$, the target rotation angular velocity $\hat{\omega}$ is multiplied with $\operatorname{sign}(i) = \left\{ \begin{matrix} +1 & i = 1 \\ -1 & i = 2 \end{matrix} \right.$ to ensure the symmetry.
The rope morphology $o^{\text{rope}, i} = \left[ \mathbf{p}^{\text{rope}, i}[k_{1}^{i}], \mathbf{p}^{\text{rope}, i}[k_{2}^{i}], \cdots, \mathbf{p}^{\text{rope}, i}[k_{m}^{i}] \right]$ concatenates positions of $m$ points within the discretely simulated rope, which are represented in the base link frame of humanoid rope turner $i$. The observing indices $k_{1:m}^{i}$ are randomly sampled at the start of the episode and ordered by their index distance to the rope end attached on humanoid rope turner $i$.
The proprioceptive sensing $o^{\text{proprio}, i} = [\boldsymbol{\omega}^{\text{b}}, \mathbf{g}^{\text{b}}, \mathbf{q}, \dot{\mathbf{q}}, \tilde{a}]$ contains base link angular velocity $\boldsymbol{\omega}^{\text{b}}$, gravity projected in the base link frame $\mathbf{g}^{\text{b}}$, joint positions $\mathbf{q}$, joint velocities $\dot{\mathbf{q}}$ and action at last timestep $\tilde{a}$.

\paragraph{Multi-Agent Policy Optimization}

Finally, with the above observations, the policy outputs action $a_{t} \in \mathbb{R}^{N_{\text{joints}}}$ to further derive the target joint positions of all joints $\hat{\mathbf{q}}_{t} = \mathbf{q}_{\text{default}} + \boldsymbol{\alpha} \odot a_{t}$ for PD controller~\cite{liao2025beyondmimic}. To ensure real-time decentralized inference, we adopt the Multi-Agent Proximal Policy Optimization (MAPPO) \cite{yu2022surprising} algorithm with Centralized Training and Decentralized Execution (CTDE) paradigm to optimize $\pi^{\text{manip}}$ under the reward terms described in Appendix.

\subsection{High-level Scheduling Policy}
\label{sec:high_level}

The aforementioned rope manipulation policy enables two humanoid robots to cooperatively control the rope. However, integrating this skill with a player to perform group long rope skipping remains unresolved. To address this issue, we further train a high-level scheduling policy $\pi^{\text{sched}}$ to coordinate with a physically controlled humanoid player.

\paragraph{Jump Rhythm Representation}
First, to obtain a reliable jumping policy, we formally model the jumping task as an alternating process between ground and air stages. To be specific, each cycle in such process can be described as a tuple $\mathcal{G}^{\text{rhythm}} = \langle \nu, \kappa, \varphi, c = \mathbf{1}(\varphi >= \kappa) \rangle$, where $\nu = 1 / T$ refers to the frequency induced by the cycle length $T$, $\kappa = T^{\text{ground}} / T$ represents the ratio of the ground stage and $\varphi \in [0, 1)$ is a continuous phase variable indicating the progress within the cycle, which steps forward via $\varphi \leftarrow \varphi + \nu \Delta t$ and is reset to 0 at the start of a new cycle.

\paragraph{Jump Policy Training}

Based on the clock signal, we train a jumping policy $\pi^{\text{jump}}(a_{t} \mid o_{t}^{\text{jump}} = [\mathcal{G}^{\text{rhythm}}, o_{t - H + 1 : t}^{\text{proprio}}])$ to provide physically plausible jumping motions, where the action $a_{t}$ and the proprioceptive sensing history $o_{t - H + 1 : t}^{\text{proprio}}$ share the same definition as $\pi^{\text{manip}}$. The core objective of policy $\pi^{\text{jump}}$ is to align the binary feet contact mask with the binary clock $c$, which can further derive the periodic jumping behavior as shown in Figure \ref{fig:main}. Additional rewards can be found in Appendix.

\paragraph{Centralized Scheduling}
With the pretrained rope manipulation policy $\pi^{\text{manip}}$, we introduce the high-level scheduling policy $\pi^{\text{sched}}(\mathcal{G}^{\text{manip}} \mid o_{t}^{\text{sched}})$ to adaptively adjust $\mathcal{G}^{\text{manip}}$.
The observation is defined as $o_{t}^{\text{sched}} = [\mathcal{G}^{\text{rhythm}}, o_{t - H + 1 : t}^{\text{rope}}, o_{t - H + 1 : t}^{\text{player}}]$, where the player state $o_{t}^{\text{player}} = [\mathbf{p}_{t}^{\text{obb}}, \mathbf{R}_{t}^{\text{obb}}, \Delta_{t}^{\text{obb}}]$ includes position $\mathbf{p}_{t}^{\text{obb}}$, orientation $\mathbf{R}_{t}^{\text{obb}}$ and size $\Delta_{t}^{\text{obb}}$ of the player Oriented Bounding Box (OBB). Both $o_{t}^{\text{rope}}$ and $o_{t}^{\text{player}}$ are represented in the rotation center frame.
The core objective of policy $\pi^{\text{sched}}$ is to synchronize with the player's jumping rhythm while reducing the collision risk between the rope and the player. For rhythm synchronization, we first define the rope rotation phase around axis $\mathbf{e}^{\text{r}}$ as
\begin{equation}
\begin{gathered}
    \bar{\theta}^{\text{rope}} = \frac{1}{2 \pi N} \sum_{i = 1}^{N} \operatorname{atan2}(\operatorname{sgn}(\langle \bar{\boldsymbol{\omega}}, \mathbf{e}^{\text{r}} \rangle) \xi_{i}^{y}, -\xi_{i}^{z}), \\
    \bar{\boldsymbol{\omega}} = \arg \min_{\boldsymbol{\omega}} \sum_{i = 1}^{N} \lVert \mathbf{v}^{\text{rope}}[i] - \boldsymbol{\omega} \times (\mathbf{p}^{\text{rope}}[i] - \mathbf{p}^{\text{c}}) \rVert_{2}^{2}, \ 
    \xi_{i} = [\mathbf{e}^{\text{r}}, \mathbf{e}^{\text{z}} \times \mathbf{e}^{\text{r}}, \mathbf{e}^{\text{z}}] (\mathbf{p}^{\text{rope}}[i] - \mathbf{p}^{\text{c}}).
\end{gathered}
\end{equation}

Ideally, the rotation phase $\bar{\theta}^{\text{rope}}$ should periodically go through 0 when the player ascends to the highest points (i.e., $\varphi = 1 - \frac{1 - \kappa}{2}$) in order to make successful rope skips. Thus, we set the target rotation phase $\hat{\theta}^{\text{rope}}$ to be $\frac{1 - \kappa}{2}$ ahead of the jump rhythm phase $\varphi$ and compute the phase tracking error along with the other reward terms listed in Appendix to optimize $\pi^{\text{sched}}$.

\subsection{Diverse Player Discovery}
\label{sec:diverse_jump}
The preceding scheduling policy $\pi^{\text{sched}}$ enables the humanoid rope turners to coordinate with a fixed-style player. However, in practice, different individuals may possess various motion patterns. To enhance the adaptability of $\pi^{\text{sched}}$ to such variations, we introduce an auxiliary intrinsic objective to discover diverse player behaviors, which are then used as counterparts in cooperative game training.

\paragraph{Diversity Intrinsic Objective}
To be specific, the jumping policy $\pi^{\text{jump}}(a_{t} \mid o_{t}^{\text{jump}}, z)$ is conditioned on a continuous latent variable $z \in \mathbb{R}^{d_{\text{latent}}}$ in addition to the original observations $o_{t}^{\text{jump}}$. To increase the discrepancy between states visited by policy with different latent $z$, we formulate a diversity intrinsic objective as follows:
\begin{equation}
    \max_{\pi^{\text{jump}}} \gamma_{\mathcal{F}} \Big( p(x, z), p(x) p(z) \Big) = \sup_{f \in \mathcal{F}} \mathbb{E}_{z \sim p(z), x \sim \pi^{\text{jump}}(z)} \Big[ f(x, z) - \mathbb{E}_{\tilde{z} \sim p(z)} f(x, \tilde{z}) \Big],
\end{equation}
where $\gamma_{\mathcal{F}}(\cdot, \cdot)$ denotes Integral Probability Metric (IPM) over the function class $\mathcal{F}$, $x \in \mathbb{R}^{d_{\text{feat}}}$ is the feature variable describing the states visited by policy, e.g. local-frame end-effectors poses.

\paragraph{Practical Implementation}
We use a standard Gaussian distribution $\mathcal{N}(\mathbf{0}, \mathbf{I}_{d_{\text{latent}}})$ as the prior latent distribution $p(z)$. We choose $\mathcal{F}$ to be the set of 1-Lipschitz continuous functions in $\mathbb{R}^{d_{\text{feat}} + d_{\text{latent}}}$, which makes $\gamma_{\mathcal{F}}(\cdot, \cdot)$ equivalent to Wasserstein-1 distance $\mathcal{I}_{\mathcal{W}}(\cdot, \cdot)$.
Since directly searching in the whole function class $\mathcal{F}$ is intractable, an extra learnable metric model $\phi : \mathbb{R}^{d_{\text{feat}} + d_{\text{latent}}} \mapsto \mathbb{R}$ is optimized along with the policy $\pi^{\text{jump}}$ to approximate the supremum in the IPM
\begin{equation}
    \max_{\phi} \mathbb{E}_{z \sim p(z), x \sim \pi^{\text{jump}}(z)} \Big[ \phi(x, z) - \mathbb{E}_{\tilde{z} \sim p(z)} \phi(x, \tilde{z}) \Big],
\end{equation}
where the Lipschitz continuity of $\phi$ is guaranteed by applying Spectral Normalization~\cite{miyato2018spectral}.
The diversity intrinsic serves as an additional reward term weighted with the task rewards, forming a composite reward $r_{t}^{\text{div}} = r_{t}^{\text{task}} + \beta \phi(x_{t}, z)$ for policy optimization. We also introduce a curriculum mechanism to enable diversity intrinsic only when the policy $\pi^{\text{jump}}$ reaches a predefined task performance threshold, thereby restricting that diversity objective does not hinder the main task objectives.


\section{Experiments}
\label{sec:result}
In this section, we conduct extensive experiments in both simulation and real-world settings to answer the following research questions:
(1) Can \MethodName perform flexible and stable rope manipulation (\cref{sec:rope_man})?
(2) Can \MethodName efficiently cooperate with diverse players (\cref{sec:coordinate})?
(3) Can \MethodName robustly transfer to real-world scenarios (\cref{sec:real_world})?

\subsection{Experimental Settings}

\paragraph{Simulation Environment}
We conduct our simulation experiments in Isaac Lab \cite{mittal2025isaaclab} for massive parallel simulation. To approximately simulate the physical properties of a rope, we create $N$ small rigid capsules and connect them via passively driven spherical joints. Each end of the rope is attached to the wrist link of the corresponding humanoid rope turner with hand removed.

\paragraph{Real-world Deployment}
We deploy \MethodName on Unitree G1 29-dof humanoid robots. An optical motion capture (MoCap) system is utilized to obtain (a) positions of $m$ reflective markers affixed to the rope (b) poses of base links in humanoid rope turners (c) positions of key body parts of the player, which are transported to each computing node for observation construction.

\begin{table}
\centering
\renewcommand\arraystretch{1.5}
\caption{\textbf{Simulation Results of Rope Manipulation}.}
\label{tab:rope_manipulation_comparison}
\resizebox{\textwidth}{!}{
\begin{tabular}{llcccc}
\toprule

& Metrics & Ours & Single Agent & Open Loop & w/o Segment Sampling \\

\midrule
\multirow{2}{*}{\textbf{Rope Manipulation}} &

$E_{\text{rot}}$ ($\downarrow$) &
$\mathbf{1.719 \pm 0.332}$ & $2.733 \pm 0.539$ & $6.754 \pm 2.018$ & $2.223 \pm 0.484$ \\

&
$E_{\text{wid}}$ ($\downarrow$) &
$\mathbf{0.063 \pm 0.011}$ & $0.136 \pm 0.030$ & $0.284 \pm 0.135$ & $0.087 \pm 0.038$ \\

\midrule
\multirow{2}{*}{\textbf{Coordinated Movement}} &

$E_{\text{lin}}$ ($\downarrow$) &
$\mathbf{0.103 \pm 0.011}$ & $0.222 \pm 0.052$ & $0.199 \pm 0.049$ & $0.121 \pm 0.024$ \\

&
$E_{\text{ang}}$ ($\downarrow$) &
$\mathbf{0.109 \pm 0.041}$ & $0.114 \pm 0.039$ & $0.112 \pm 0.040$ & $0.110 \pm 0.041$ \\

\midrule
\multirow{2}{*}{\textbf{Control Stability}} &

Act. Rate ($\downarrow$) &
$1.349 \pm 0.130$ & $1.526 \pm 0.207$ & $\mathbf{0.945 \pm 0.097}$ & $1.300 \pm 0.142$ \\

&
Feet Slip. ($\downarrow$) &
$\mathbf{0.040 \pm 0.003}$ & $0.122 \pm 0.010$ & $0.082 \pm 0.012$ & $0.053 \pm 0.004$ \\

\bottomrule
\end{tabular}
}
\end{table}

\subsection{Rope Manipulation Analysis}
\label{sec:rope_man}
\paragraph{Baselines} We first evaluate the effectiveness of the rope manipulation component of \MethodName and compare it with several baselines defined as follows: (1) \textbf{Single Agent} uses a single-agent policy to control two humanoids simultaneously, where the observation input and action output are formed by concatenating the observations and actions of our policy, i.e., $[o_t^{\text{manip},1}, o_t^{\text{manip},2}]$ and $[a_t^1, a_t^2]$; (2) \textbf{Open Loop} records the evaluation trajectories of our policy and then reproduces the motion independently on each humanoid using a motion tracking method~\cite{liao2025beyondmimic}; while (3) \textbf{w/o Segment Sampling} uses uniformly spaced rope observation indices instead of randomly sampled ones during training.

\paragraph{Metrics} In the comparison, we evaluate the performance of rope manipulation using the following metrics: (1) \textbf{Rotation Tracking Error} $E_{\text{rot}}=\mathbb{E} [\lVert \bar{\boldsymbol{\omega}} - \hat{\boldsymbol{\omega}} \rVert_{2}]$; (2) \textbf{Width Tracking Error} $E_{\text{wid}}=\mathbb{E}[ \lvert w - \hat{w} \rvert]$; (3) \textbf{Linear Velocity Tracking Error} $E_{\text{lin}}=\mathbb{E}[ \lVert v_{xy}^{\text{lin}} - \hat{v}_{xy}^{\text{lin}} \rVert_{2}]$; (4) \textbf{Angular Velocity Tracking Error} $E_{\text{ang}}=\mathbb{E}[ \lvert v_{z}^{\text{ang}} - \hat{v}_{z}^{\text{ang}} \rvert]$; (5) \textbf{Action Rate}, which measures the difference between actions at adjacent timesteps; and (6) \textbf{Feet Slippage}, defined as the horizontal sliding velocity of the feet when they are in contact with the ground. Metrics (1)–(4) evaluate task completion performance, while (5)–(6) measure the control stability. All metrics are estimated over 1,000 episodes.

\paragraph{Comparison Results} The experimental results are presented in \cref{tab:rope_manipulation_comparison}. First, Single Agent performs consistently worse than our method across all metrics, suggesting that modeling the task as a single-agent control problem introduces redundant and highly coupled observation-action spaces, thereby reducing optimization efficiency. Second, Open Loop achieves slightly smoother actions, but its rope manipulation accuracy drops substantially, highlighting the limitations of blind motion tracking in handling the complex deformable object dynamics involved in rope turning and demonstrates the necessity of closed-loop reinforcement learning. Finally, w/o Segment Sampling shows mild degradation in both rope manipulation and coordinated-movement metrics, indicating weaker robustness to disturbances from randomly spaced rope observation points, which more closely resemble real-world sensing conditions. Taken together, these results validate the effectiveness of our module designs for low-level rope manipulation.

\subsection{Player Coordination Analysis}
\label{sec:coordinate}

\begin{table}[t]
\centering
\caption{\textbf{Simulation Results of Player Coordination}.}
\label{tab:player_coordination_comparision}
\begin{tabular}{lccc}
\toprule
Metrics & Ours & w/o Scheduling & w/o Player Diversity \\
\midrule
Overlap Ratio ($\downarrow$) &
$\mathbf{0.090 \pm 0.049}$ &
$0.392 \pm 0.140$ &
$0.120 \pm 0.094$ \\

Phase Tracking Error ($\downarrow$) &
$\mathbf{0.397 \pm 0.115}$ &
$1.092 \pm 0.273$ &
$0.414 \pm 0.155$ \\

Player Tracking Error ($\downarrow$) &
$\mathbf{0.116 \pm 0.022}$ &
$0.182 \pm 0.360$ &
$0.172 \pm 0.021$ \\

Complete Rate ($\uparrow$) &
$\mathbf{0.787 \pm 0.204}$ &
$0.358 \pm 0.207$ &
$0.751 \pm 0.216$ \\
\bottomrule
\end{tabular}
\end{table}

\paragraph{Baselines and Metrics} We then evaluate the cooperative rope skipping performance of \MethodName and the baseline methods with a diverse jumping policy conditioned on randomly sampled latent codes in simulation. The baselines include (1) \textbf{w/o Scheduling}, which disables the high-level scheduling policy and instead uses the clock frequency to compute the target rotational angular velocity for in-place rope turning, and (2) \textbf{w/o Player Diversity}, which adopts another jumping policy trained without diversity intrinsic as counterparts during training. We use the following metrics for evaluation: (1) \textbf{Overlap Ratio}, measured as the fraction of timesteps in which the rope overlaps with the player’s OBB; (2) \textbf{Phase Tracking Error}, measured as the discrepancy between the rope rotation phase $\bar{\theta}^{\text{rope}}$ and the derived target rotation phase $\hat{\theta}^{\text{rope}}$; (3) \textbf{Player Tracking Error}, measured as the horizontal distance between the rotation center and the center of player’s OBB; and (4) \textbf{Complete Rate}, measured as the percentage of successful jumps among all attempts.

\paragraph{Comparison Results}
The comparison results are presented in \cref{tab:player_coordination_comparision}. w/o Scheduling shows a significant performance degradation, suggesting that a simple rule-based combination cannot produce the complex cooperative behavior required for synchronized long rope skipping, where precise coordination is essential, highlighting the necessity of the proposed adaptive high-level scheduling policy. In addition, w/o Player Diversity similarly underperforms our method, indicating that introducing diverse counterparts during training is crucial for improving the generalization and adaptability of the learned coordination policy.

\begin{figure}
    \centering

    \begin{subfigure}{0.27\textwidth}
        \centering
        \includegraphics[width=\textwidth]{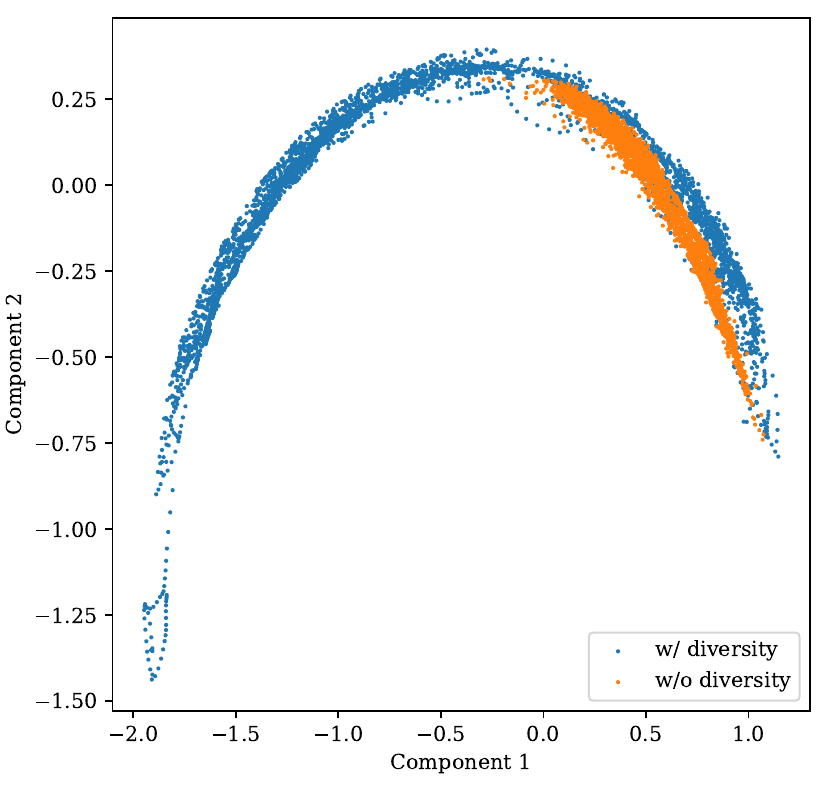}
        \caption{}
        \label{fig:ee_pose_visualization}
    \end{subfigure}
    \hfill
    \begin{subfigure}{0.72\textwidth}
        \centering
        \includegraphics[width=\linewidth]{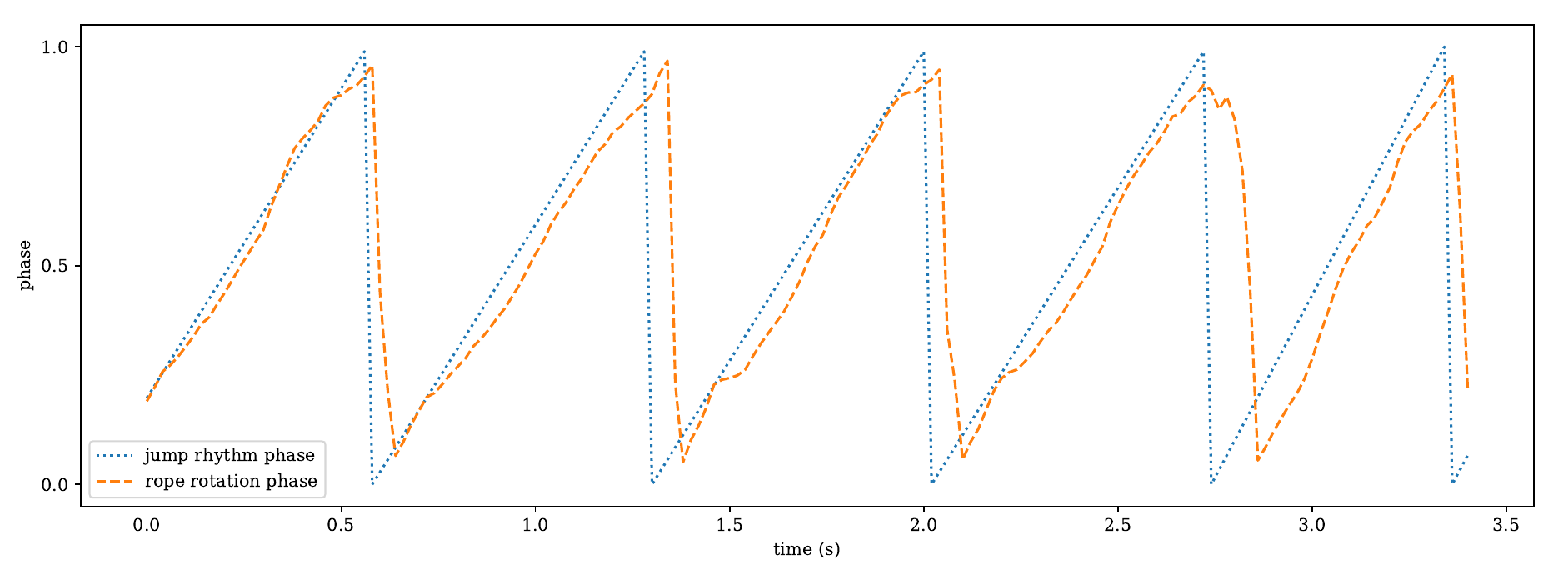}
        \caption{}
        \label{fig:rhythm_align}
    \end{subfigure}

    \caption{\textbf{Visualization Results}. (a) PCA-embedding of visited local-frame end-effector poses. (b) Plot of offset rope rotation phase and jump rhythm phase over time.}
\end{figure}

\paragraph{Player Diversity Visualization}
We further use PCA to visualize the local-frame end-effector poses visited by jumping policy with and without the diversity intrinsic objective, as shown in \cref{fig:ee_pose_visualization}. The results show that introducing the diversity objective substantially expands the behavior distribution of the jumping policy, demonstrating its effectiveness in discovering diverse player behaviors.

\paragraph{Phase Tracking Visualization}
Finally, we visualize the rope rotation phase $\bar{\theta}^{\text{rope}}$ offset by $\frac{1 - \kappa}{2}$ together with the corresponding jump rhythm phase $\varphi$ in \cref{fig:rhythm_align}. The results indicate that two phases can stay synchronized within a small tracking error, further demonstrating the ability of the high-level scheduling policy to effectively adapt to the jumping policy.


\subsection{Real-world Deployment}
\label{sec:real_world}
\begin{figure}
	\centering
	\includegraphics[width=\linewidth]{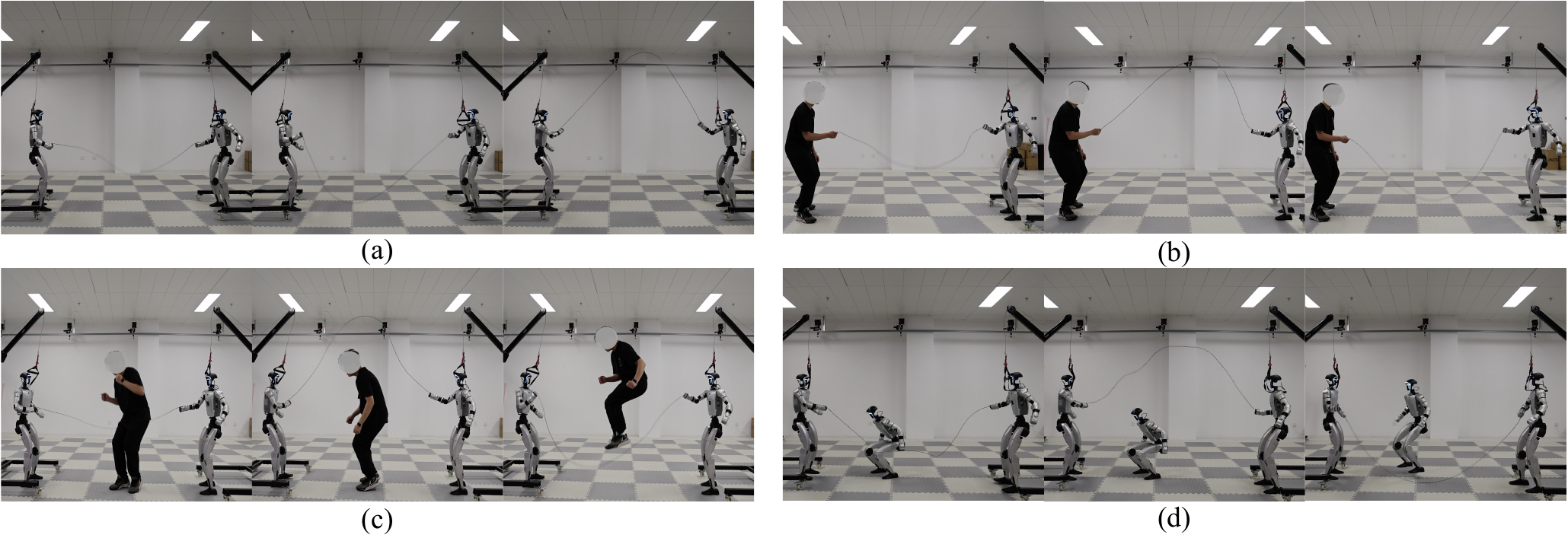}
    \caption{\textbf{Real-world Deployment}. (a) Humanoid-humanoid rope turning. (b) Humanoid-human rope turning. (c) Humanoids rope turning for a human. (d) Humanoids rope turning for a humanoid.}
    \label{fig:deploy}
\end{figure}

In this section, we deploy our method in four real-world scenarios listed in Figure \ref{fig:deploy}. In the pure rope turning setting, our method enables two humanoids to coordinate with each other, while also allowing a single humanoid to cooperate with a human partner, demonstrating the robustness of \MethodName's rope manipulation. Furthermore, when a jumping player is involved, our method successfully coordinates with both human and humanoid players, despite their large differences in jumping styles and body morphology, further confirming the strong adaptability of \MethodName to diverse players.

\section{Conclusion}
\label{sec:conclusion}
We propose \MethodName, a hierarchical RL framework for cooperative long rope skipping with multiple humanoid robots. \MethodName first learns a low-level decentralized rope manipulation policy using MAPPO. It then introduces a centralized high-level scheduling policy to follow the player’s rhythm and prevent collisions between the rope and the player. To further enhance adaptability, \MethodName augments the training process of high-level scheduling policy with automatically discovered diverse player behavior, improving generalization to different jumping styles. Extensive simulation and real-world experiments show that \MethodName achieves stable and efficient rope manipulation and can robustly cooperate with various players.

\section{Limitations}
One limitation of this work is that the proposed framework is tailored to the specific task of long rope skipping, and thus cannot be directly generalized to other cooperative humanoid tasks. Moreover, we currently focus only on a simple single-player setting, while extending the framework to multiple players and more challenging fancy rope skipping techniques (e.g. double dutch) remains an open problem. Finally, enabling humanoids to collaborate with diverse human partners in the wild via on-board sensors rather than MoCap system represents a promising direction for future work.

\clearpage
\newpage


\bibliographystyle{ACM-Reference-Format}
\bibliography{reference}

@inproceedings{miyato2018spectral,
  title={Spectral Normalization for Generative Adversarial Networks},
  author={Miyato, Takeru and Kataoka, Toshiki and Koyama, Masanori and Yoshida, Yuichi},
  booktitle={International Conference on Learning Representations},
  year={2018}
}

@inproceedings{hirai1998development,
  title={The development of Honda humanoid robot},
  author={Hirai, Kazuo and Hirose, Masato and Haikawa, Yuji and Takenaka, Toru},
  booktitle={Proceedings. 1998 IEEE international conference on robotics and automation},
  volume={2},
  pages={1321--1326},
  year={1998},
  organization={IEEE}
}

@article{tong2024advancements,
  title={Advancements in humanoid robots: A comprehensive review and future prospects},
  author={Tong, Yuchuang and Liu, Haotian and Zhang, Zhengtao},
  journal={IEEE/CAA Journal of Automatica Sinica},
  volume={11},
  number={2},
  pages={301--328},
  year={2024},
  publisher={IEEE/CAA Journal of Automatica Sinica}
}

@inproceedings{he2024learning,
  title={Learning human-to-humanoid real-time whole-body teleoperation},
  author={He, Tairan and Luo, Zhengyi and Xiao, Wenli and Zhang, Chong and Kitani, Kris and Liu, Changliu and Shi, Guanya},
  booktitle={2024 IEEE/RSJ International Conference on Intelligent Robots and Systems},
  pages={8944--8951},
  year={2024},
  organization={IEEE}
}

@inproceedings{crowley2023optimizing,
  title={Optimizing bipedal locomotion for the 100m dash with comparison to human running},
  author={Crowley, Devin and Dao, Jeremy and Duan, Helei and Green, Kevin and Hurst, Jonathan and Fern, Alan},
  booktitle={2023 IEEE International Conference on Robotics and Automation},
  pages={12205--12211},
  year={2023},
  organization={IEEE}
}

@article{qi2023vertical,
  title={Vertical jump of a humanoid robot with cop-guided angular momentum control and impact absorption},
  author={Qi, Haoxiang and Chen, Xuechao and Yu, Zhangguo and Huang, Gao and Liu, Yaliang and Meng, Libo and Huang, Qiang},
  journal={IEEE Transactions on Robotics},
  volume={39},
  number={4},
  pages={3154--3166},
  year={2023},
  publisher={IEEE}
}

@article{qin2018music,
  title={A music-driven dance system of humanoid robots},
  author={Qin, Ruilin and Zhou, Changle and Zhu, He and Shi, Minghui and Chao, Fei and Li, Na},
  journal={International Journal of Humanoid Robotics},
  volume={15},
  number={05},
  pages={1850023},
  year={2018},
  publisher={World Scientific}
}

@inproceedings{zhuang2025humanoid,
  title={Humanoid Parkour Learning},
  author={Zhuang, Ziwen and Yao, Shenzhe and Zhao, Hang},
  booktitle={Conference on Robot Learning},
  pages={1975--1991},
  year={2025},
  organization={PMLR}
}

@article{su2025hitter,
  title={Hitter: A humanoid table tennis robot via hierarchical planning and learning},
  author={Su, Zhi and Zhang, Bike and Rahmanian, Nima and Gao, Yuman and Liao, Qiayuan and Regan, Caitlin and Sreenath, Koushil and Sastry, S Shankar},
  journal={arXiv preprint arXiv:2508.21043},
  year={2025}
}

@inproceedings{li2019using,
  title={Using deep reinforcement learning to learn high-level policies on the atrias biped},
  author={Li, Tianyu and Geyer, Hartmut and Atkeson, Christopher G and Rai, Akshara},
  booktitle={2019 International Conference on Robotics and Automation},
  pages={263--269},
  year={2019},
  organization={IEEE}
}

@inproceedings{li2021reinforcement,
  title={Reinforcement learning for robust parameterized locomotion control of bipedal robots},
  author={Li, Zhongyu and Cheng, Xuxin and Peng, Xue Bin and Abbeel, Pieter and Levine, Sergey and Berseth, Glen and Sreenath, Koushil},
  booktitle={2021 IEEE International Conference on Robotics and Automation},
  pages={2811--2817},
  year={2021},
  organization={IEEE}
}

@article{radosavovic2024real,
  title={Real-world humanoid locomotion with reinforcement learning},
  author={Radosavovic, Ilija and Xiao, Tete and Zhang, Bike and Darrell, Trevor and Malik, Jitendra and Sreenath, Koushil},
  journal={Science Robotics},
  volume={9},
  number={89},
  pages={eadi9579},
  year={2024},
  publisher={American Association for the Advancement of Science}
}

@inproceedings{radosavovic2024humanoid,
  title={Humanoid locomotion as next token prediction},
  author={Radosavovic, Ilija and Zhang, Bike and Shi, Baifeng and Rajasegaran, Jathushan and Kamat, Sarthak and Darrell, Trevor and Sreenath, Koushil and Malik, Jitendra},
  booktitle={Proceedings of the 38th International Conference on Neural Information Processing Systems},
  pages={79307--79324},
  year={2024}
}

@inproceedings{long2025learning,
  title={Learning humanoid locomotion with perceptive internal model},
  author={Long, Junfeng and Ren, Junli and Shi, Moji and Wang, Zirui and Huang, Tao and Luo, Ping and Pang, Jiangmiao},
  booktitle={2025 IEEE International Conference on Robotics and Automation},
  pages={9997--10003},
  year={2025},
  organization={IEEE}
}

@article{cheng2024expressive,
  title={Expressive whole-body control for humanoid robots},
  author={Cheng, Xuxin and Ji, Yandong and Chen, Junming and Yang, Ruihan and Yang, Ge and Wang, Xiaolong},
  journal={arXiv preprint arXiv:2402.16796},
  year={2024}
}

@inproceedings{ji2025exbody2,
  title={ExBody2: Advanced Expressive Humanoid Whole-Body Control},
  author={Ji, Mazeyu and Peng, Xuanbin and Liu, Fangchen and Li, Jialong and Yang, Ge and Cheng, Xuxin and Wang, Xiaolong},
  booktitle={RSS 2025 Workshop on Whole-body Control and Bimanual Manipulation: Applications in Humanoids and Beyond},
  year = {2025}
}

@article{chen2026learning,
  title={Learning human-like badminton skills for humanoid robots},
  author={Chen, Yeke and Dong, Shihao and Ji, Xiaoyu and Sun, Jingkai and Luo, Zeren and Zhao, Liu and Zhang, Jiahui and Li, Wanyue and Ma, Ji and Xu, Bowen and others},
  journal={arXiv preprint arXiv:2602.08370},
  year={2026}
}

@article{sunehag2017value-vdn,
  title={Value-decomposition networks for cooperative multi-agent learning},
  author={Sunehag, Peter and Lever, Guy and Gruslys, Audrunas and Czarnecki, Wojciech Marian and Zambaldi, Vinicius and Jaderberg, Max and Lanctot, Marc and Sonnerat, Nicolas and Leibo, Joel Z and Tuyls, Karl and others},
  journal={arXiv preprint arXiv:1706.05296},
  year={2017}
}

@inproceedings{rashid2018qmix,
  title={QMIX: Monotonic Value Function Factorisation for Deep Multi-Agent Reinforcement Learning},
  author={Rashid, Tabish and Samvelyan, Mikayel and Schroeder, Christian and Farquhar, Gregory and Foerster, Jakob and Whiteson, Shimon},
  booktitle={International Conference on Machine Learning},
  pages={4295--4304},
  year={2018},
}

@inproceedings{wang2021dop,
title={{DOP}: Off-Policy Multi-Agent Decomposed Policy Gradients},
author={Yihan Wang and Beining Han and Tonghan Wang and Heng Dong and Chongjie Zhang},
booktitle={International Conference on Learning Representations},
year={2021},
}

@article{feng2026multi,
  title={Multi-agent embodied ai: Advances and future directions},
  author={Feng, Zhaohan and Xue, Ruiqi and Yuan, Lei and Yu, Yang and Ding, Ning and Liu, Meiqin and Gao, Bingzhao and Sun, Jian and Zheng, Xinhu and Wang, Gang},
  journal={Science China Information Sciences},
  volume={69},
  number={5},
  pages={151202},
  year={2026},
  publisher={Springer}
}

@inproceedings{lowe2017multi,
  title={Multi-agent actor-critic for mixed cooperative-competitive environments},
  author={Lowe, Ryan and Wu, Yi and Tamar, Aviv and Harb, Jean and Abbeel, Pieter and Mordatch, Igor},
  booktitle={Proceedings of the 31st International Conference on Neural Information Processing Systems},
  pages={6382--6393},
  year={2017}
}

@inproceedings{yu2022surprising,
  title={The surprising effectiveness of PPO in cooperative multi-agent games},
  author={Yu, Chao and Velu, Akash and Vinitsky, Eugene and Gao, Jiaxuan and Wang, Yu and Bayen, Alexandre and Wu, Yi},
  booktitle={Proceedings of the 36th International Conference on Neural Information Processing Systems},
  pages={24611--24624},
  year={2022}
}

@article{zhang2024multi,
  title={Multi-agent reinforcement learning for autonomous driving: A survey},
  author={Zhang, Ruiqi and Hou, Jing and Walter, Florian and Gu, Shangding and Guan, Jiayi and R{\"o}hrbein, Florian and Du, Yali and Cai, Panpan and Chen, Guang and Knoll, Alois},
  journal={arXiv preprint arXiv:2408.09675},
  year={2024}
}

@inproceedings{fang2023learning,
  title={Learning multi-agent intention-aware communication for optimal multi-order execution in finance},
  author={Fang, Yuchen and Tang, Zhenggang and Ren, Kan and Liu, Weiqing and Zhao, Li and Bian, Jiang and Li, Dongsheng and Zhang, Weinan and Yu, Yong and Liu, Tie-Yan},
  booktitle={Proceedings of the 29th ACM SIGKDD Conference on Knowledge Discovery and Data Mining},
  pages={4003--4012},
  year={2023}
}

@article{zhu2022survey,
  title={A Survey of Multi-Agent Reinforcement Learning with Communication},
  author={Zhu, Changxi and Dastani, Mehdi and Wang, Shihan},
  journal={preprint arXiv:2203.08975},
  year={2022}
}

@inproceedings{zhang2023discovering,
  title={Discovering generalizable multi-agent coordination skills from multi-task offline data},
  author={Zhang, Fuxiang and Jia, Chengxing and Li, Yi-Chen and Yuan, Lei and Yu, Yang and Zhang, Zongzhang},
  booktitle={The Eleventh International Conference on Learning Representations},
  year={2023}
}

@article{wang2022model,
  title={Model-Based Multi-Agent Reinforcement Learning: Recent Progress and Prospects},
  author={Wang, Xihuai and Zhang, Zhicheng and Zhang, Weinan},
  journal={preprint arXiv:2203.10603},
  year={2022}
}

@article{guo2022towards,
  title={Towards Comprehensive Testing on the Robustness of Cooperative Multi-Agent Reinforcement Learning},
  author={Guo, Jun and Chen, Yonghong and Hao, Yihang and Yin, Zixin and Yu, Yin and Li, Simin},
  journal={preprint arXiv:2204.07932},
  year={2022}
}

@article{liao2025beyondmimic,
  title={Beyondmimic: From motion tracking to versatile humanoid control via guided diffusion},
  author={Liao, Qiayuan and Truong, Takara E and Huang, Xiaoyu and Gao, Yuman and Tevet, Guy and Sreenath, Koushil and Liu, C Karen},
  journal={arXiv preprint arXiv:2508.08241},
  year={2025}
}

@article{mittal2025isaaclab,
  title={Isaac Lab: A GPU-Accelerated Simulation Framework for Multi-Modal Robot Learning},
  author={Mayank Mittal and Pascal Roth and James Tigue and Antoine Richard and Octi Zhang and Peter Du and Antonio Serrano-Muñoz and Xinjie Yao and René Zurbrügg and Nikita Rudin and Lukasz Wawrzyniak and Milad Rakhsha and Alain Denzler and Eric Heiden and Ales Borovicka and Ossama Ahmed and Iretiayo Akinola and Abrar Anwar and Mark T. Carlson and Ji Yuan Feng and Animesh Garg and Renato Gasoto and Lionel Gulich and Yijie Guo and M. Gussert and Alex Hansen and Mihir Kulkarni and Chenran Li and Wei Liu and Viktor Makoviychuk and Grzegorz Malczyk and Hammad Mazhar and Masoud Moghani and Adithyavairavan Murali and Michael Noseworthy and Alexander Poddubny and Nathan Ratliff and Welf Rehberg and Clemens Schwarke and Ritvik Singh and James Latham Smith and Bingjie Tang and Ruchik Thaker and Matthew Trepte and Karl Van Wyk and Fangzhou Yu and Alex Millane and Vikram Ramasamy and Remo Steiner and Sangeeta Subramanian and Clemens Volk and CY Chen and Neel Jawale and Ashwin Varghese Kuruttukulam and Michael A. Lin and Ajay Mandlekar and Karsten Patzwaldt and John Welsh and Huihua Zhao and Fatima Anes and Jean-Francois Lafleche and Nicolas Moënne-Loccoz and Soowan Park and Rob Stepinski and Dirk Van Gelder and Chris Amevor and Jan Carius and Jumyung Chang and Anka He Chen and Pablo de Heras Ciechomski and Gilles Daviet and Mohammad Mohajerani and Julia von Muralt and Viktor Reutskyy and Michael Sauter and Simon Schirm and Eric L. Shi and Pierre Terdiman and Kenny Vilella and Tobias Widmer and Gordon Yeoman and Tiffany Chen and Sergey Grizan and Cathy Li and Lotus Li and Connor Smith and Rafael Wiltz and Kostas Alexis and Yan Chang and David Chu and Linxi "Jim" Fan and Farbod Farshidian and Ankur Handa and Spencer Huang and Marco Hutter and Yashraj Narang and Soha Pouya and Shiwei Sheng and Yuke Zhu and Miles Macklin and Adam Moravanszky and Philipp Reist and Yunrong Guo and David Hoeller and Gavriel State},
  journal={arXiv preprint arXiv:2511.04831},
  year={2025},
  url={https://arxiv.org/abs/2511.04831}
}

@article{liu2023robotic,
  title={Robotic manipulation of deformable rope-like objects using differentiable compliant position-based dynamics},
  author={Liu, Fei and Su, Entong and Lu, Jingpei and Li, Mingen and Yip, Michael C},
  journal={IEEE Robotics and Automation Letters},
  volume={8},
  number={7},
  pages={3964--3971},
  year={2023},
  publisher={IEEE}
}

@article{bruce2017one,
  title={One-shot reinforcement learning for robot navigation with interactive replay},
  author={Bruce, Jake and S{\"u}nderhauf, Niko and Mirowski, Piotr and Hadsell, Raia and Milford, Michael},
  journal={arXiv preprint arXiv:1711.10137},
  year={2017}
}

@article{di2024effectiveness,
  title={On the effectiveness of retrieval, alignment, and replay in manipulation},
  author={Di Palo, Norman and Johns, Edward},
  journal={IEEE Robotics and Automation Letters},
  volume={9},
  number={3},
  pages={2032--2039},
  year={2024},
  publisher={IEEE}
}

@article{thuruthel2018stable,
  title={Stable open loop control of soft robotic manipulators},
  author={Thuruthel, Thomas George and Falotico, Egidio and Manti, Mariangela and Laschi, Cecilia},
  journal={IEEE Robotics and Automation Letters},
  volume={3},
  number={2},
  pages={1292--1298},
  year={2018},
  publisher={IEEE}
}

@article{yuan2023survey,
  title={A survey of progress on cooperative multi-agent reinforcement learning in open environment},
  author={Yuan, Lei and Zhang, Ziqian and Li, Lihe and Guan, Cong and Yu, Yang},
  journal={arXiv preprint arXiv:2312.01058},
  year={2023}
}

@article{sferrazza2024humanoidbench,
  title={Humanoidbench: Simulated humanoid benchmark for whole-body locomotion and manipulation},
  author={Sferrazza, Carmelo and Huang, Dun-Ming and Lin, Xingyu and Lee, Youngwoon and Abbeel, Pieter},
  journal={arXiv preprint arXiv:2403.10506},
  year={2024}
}

@inproceedings{he2025omnih2o,
  title={OmniH2O: Universal and Dexterous Human-to-Humanoid Whole-Body Teleoperation and Learning},
  author={He, Tairan and Luo, Zhengyi and He, Xialin and Xiao, Wenli and Zhang, Chong and Zhang, Weinan and Kitani, Kris M and Liu, Changliu and Shi, Guanya},
  booktitle={Conference on Robot Learning},
  pages={1516--1540},
  year={2025},
  organization={PMLR}
}

@article{zhang2026learning,
  title={Learning athletic humanoid tennis skills from imperfect human motion data},
  author={Zhang, Zhikai and Lu, Haofei and Lian, Yunrui and Chen, Ziqing and Liu, Yun and Lin, Chenghuai and Xue, Han and Zeng, Zicheng and Qi, Zekun and Zheng, Shaolin and others},
  journal={arXiv preprint arXiv:2603.12686},
  year={2026}
}

\newpage
\appendix
\begin{algorithm}[t]
\caption{Training Procedure of \MethodName}
\label{alg:training_procedure}
\begin{algorithmic}[1]

\REQUIRE Simulation Environments $\mathcal{E}^{\text{jump}}$, $\mathcal{E}^{\text{manip}}$, $\mathcal{E}^{\text{sched}}$
\ENSURE Player Policy $\pi^{\text{jump}}$, Rope Manipulation Policy $\pi^{\text{manip}}$, Scheduling Policy $\pi^{\text{sched}}$

\STATE \textbf{Stage I: Diverse Player Discovery}
\STATE Initialize jumping policy $\pi^{\text{jump}}$ and metric model $\phi$
\FOR{each training iteration}
    \STATE Roll out $\pi^{\text{jump}}$ in $\mathcal{E}^{\text{jump}}$
    \STATE Construct positive pairs $\mathcal{B}^{+} = \{(x_{i}, z_{i})\}$
    \STATE Sample $\tilde{z}_{i} \sim p(z)$ and construct negative pairs $\mathcal{B}^{-} = \{(x_{i}, \tilde{z}_{i})\}$
    \STATE Update $\phi$ to maximize
    \[
        \frac{1}{| \mathcal{B}^{+} |} \sum_{(x, z) \in \mathcal{B}^{+}} \phi(x, z) - \frac{1}{| \mathcal{B}^{-} |} \sum_{(x, \tilde{z}) \in \mathcal{B}^{-}} \phi(x, \tilde{z})
    \]
    \STATE Use PPO to optimize $\pi^{\text{jump}}$ with rewards in Table \ref{tab:reward_terms} and diversity intrinsic $\phi(x_{i}, z_{i})$
\ENDFOR

\STATE
\STATE \textbf{Stage II: Decentralized Cooperative Rope Manipulation}
\STATE Initialize rope manipulation policy $\pi^{\text{manip}}_{\phi}$ with parameter sharing
\FOR{each training iteration}
    \STATE Roll out $\pi^{\text{manip}}$ in $\mathcal{E}^{\text{manip}}$
    \STATE Use MAPPO to optimize $\pi^{\text{manip}}_{\phi}$ with rewards in Table \ref{tab:reward_terms}
\ENDFOR

\STATE
\STATE \textbf{Stage III: High-level Scheduling}
\STATE Load and freeze $\pi^{\text{jump}}$ and $\pi^{\text{manip}}$
\STATE Initialize scheduling policy $\pi^{\text{sched}}$
\FOR{each training iteration}
    \STATE Roll out $\pi^{\text{sched}}$ in $\mathcal{E}^{\text{sched}}$ with $\pi^{\text{jump}}$ and $\pi^{\text{manip}}$
    \STATE Use PPO to optimize $\pi^{\text{sched}}$ with rewards in Table \ref{tab:reward_terms}
\ENDFOR
\RETURN $\pi^{\text{jump}}, \pi^{\text{manip}}, \pi^{\text{sched}}$
\end{algorithmic}
\end{algorithm}

\section{Experimental Details}

\begin{figure}[htb]
	\centering
	\includegraphics[width=\linewidth]{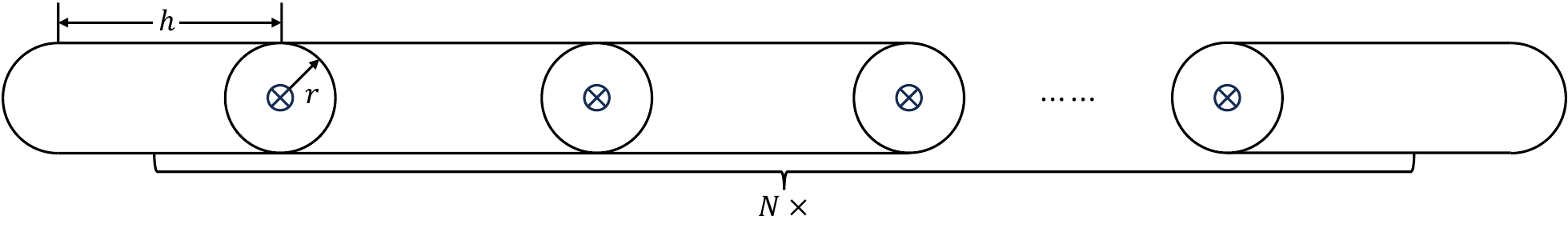}
	\caption{Schematic diagram of rope built in simulation environments.}
    \label{fig:sim_rope}
\end{figure}

\begin{figure}[htb]
    \begin{minipage}{0.6\textwidth}
        \centering
        \begin{tabular}{ll}
            \toprule
            \textbf{Parameter} & \textbf{Value} \\
            \midrule
            Capsule Height ($h$) & 0.03 m \\
            Capsule Radius ($r$) & 0.003 m \\
            Capsule density & $1.1\ \text{g} / \text{cm}^{3}$ \\
            Bend angle limit & $120^{\circ}$ \\
            Twist angle limit & $30^{\circ}$ \\
            Attach offset to wrist link & 0.08 m \\
            \bottomrule
        \end{tabular}
        \captionof{table}{Default simulation parameters of rope.}
        \label{tab:default_rope_sim_params}
    \end{minipage}
    \hfill
    \begin{minipage}{0.4\textwidth}
        \centering
    	\includegraphics[width=\linewidth]{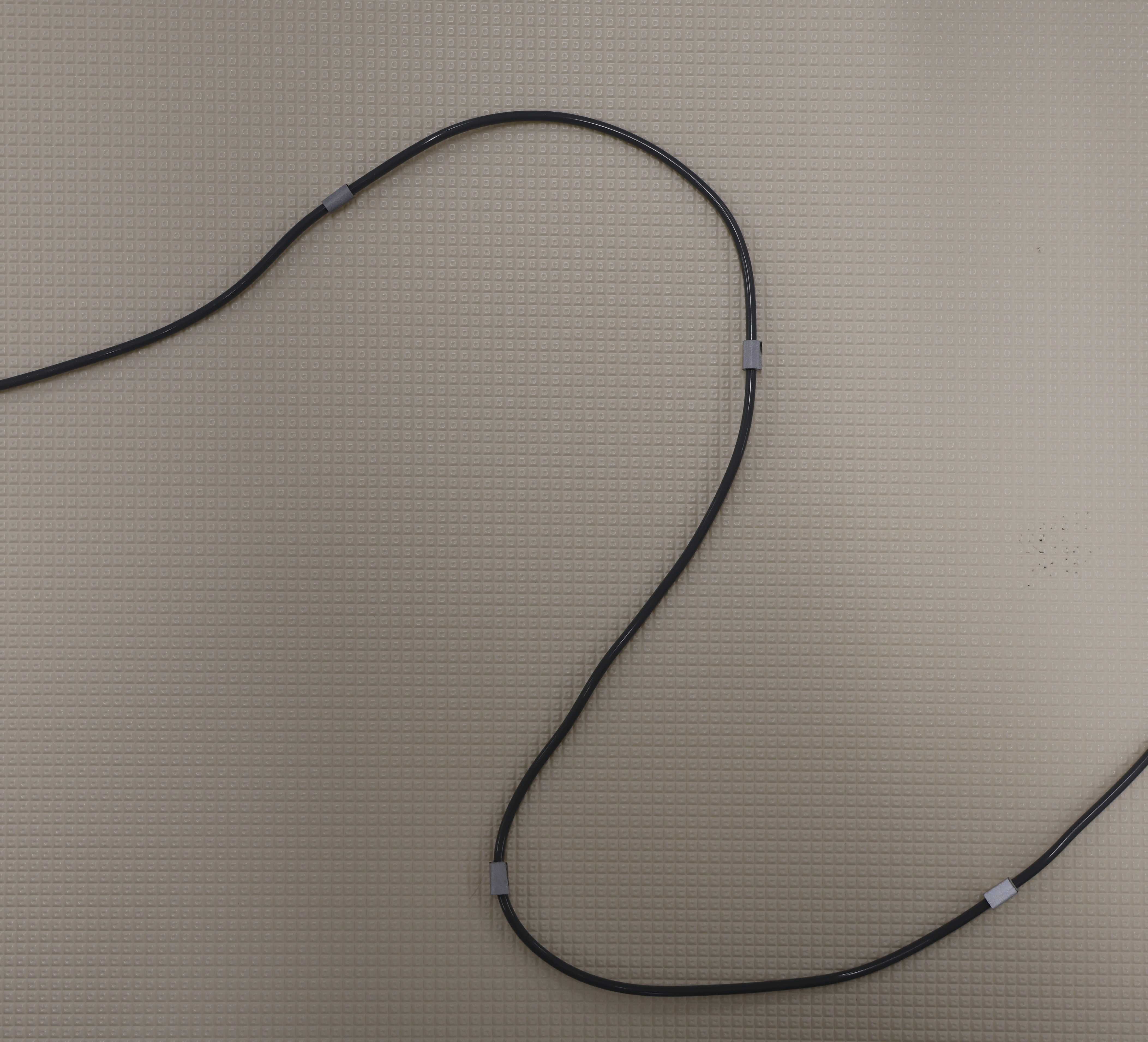}
        \caption{Rope with reflective markers.}
        \label{fig:real_rope}
    \end{minipage}
\end{figure}

\paragraph{Simulation Environment}
As illustrated in Section \ref{sec:result} and Figure \ref{fig:sim_rope}, we approximate the rope as a uniform lumped multi-body system, where $N$ rigid capsules are connected by D6 joints in Physx, the physics SDK used in Isaac Lab~\cite{mittal2025isaaclab}. For each joint, we lock three translation DoFs to enforce local inextensibility and retain the rest three rotational DoFs to allow bending and twisting, which are applied with stiffness and damping drive properties to provide proportional passive torques. Furthermore, we assume the rope to be homogeneous and transversely isotropic. Therefore, all rigid capsules are set with the same density, all internal joints as well as two bending axes in each joint are set with the same drive properties. The default simulation parameters for rope are shown in Table \ref{tab:default_rope_sim_params}.

\paragraph{Real-world Deployment}
To construct the observations of rope morphology $o^{\text{rope}}$, we affix 4 reflective markers on each side of the rope as shown in Figure \ref{fig:real_rope}, which forms 8 observable points in total to keep consistent with the design in Table \ref{tab:ppo_hyperparameters_low_level}. Similarly, when coordinating with players, we place reflective markers on players to first obtain the positions of key body parts and then compute the yaw-only OBB to form $o^{\text{player}}$. The rhythm signals $\mathcal{G}^{\text{rhythm}}$ are managed by a control terminal and transported to each humanoid robot along with the MoCap information through DDS services.

\section{Sim2Real Transfer}

\begin{table}[htb]
\footnotesize
\setlength{\tabcolsep}{4pt}
\renewcommand{\arraystretch}{1.12}

\caption{Sim2Real transfer strategies in \MethodName.}
\centering

\resizebox{\textwidth}{!}{
\begin{tabular}{@{} ll @{\hspace{30pt}} l @{}}
\toprule
 & \textbf{Term} & \textbf{Value}\\
\midrule

\multirow{13}{2.8cm}{\textbf{Dynamics\\Randomization}}
& \multicolumn{2}{@{}l}{\textit{Rope}} \\
& Number of rigid capsules ($N$)
  & $\mathcal{U}(80, 100)$ \\
& Capsule density
  & $\mathcal{U}(0.8, 1.2) \times \text{default}$\\
& Bend stiffness coefficient
  & $\mathcal{U}(10.0, 40.0)$\\
& Bend damping coefficient
  & $\mathcal{U}(2.0, 10.0)$\\
& Twist stiffness coefficient
  & $\mathcal{U}(4.0, 16.0)$\\
& Twist damping coefficient
  & $\mathcal{U}(1.0, 5.0)$\\

& \multicolumn{2}{@{}l}{\textit{Humanoid}} \\
& Static friction coefficient
  & $\mathcal{U}(0.3, 1.6)$\\
& Dynamic friction coefficient
  & $\mathcal{U}(0.3, 1.2)$\\
& Restitution coefficient
  & $\mathcal{U}(0.0, 0.5)$\\
& Body link mass
  & $\mathcal{U}(0.9, 1.1) \times \text{default}$\\
& Torso CoM offset
  & $\mathcal{U}([-0.025, 0.025], [-0.05, 0.05], [-0.05, 0.05])\ \text{m}$\\
& Push robot velocities
  & $v_x, v_y \in \mathcal{U}(-0.5, 0.5)\ \text{m} / \text{s}$ \\
& Push robot interval
  & $\mathcal{U}(1.0, 3.0)\ \text{s}$ \\
\midrule

\multirow{11}{2.8cm}{\textbf{Observations\\Noise}}
& \multicolumn{2}{@{}l}{\textit{Propiroception}} \\
& Projected gravity
  & $\mathcal{U}(-0.05, 0.05)$\\
& Base angular velocity
  & $\mathcal{U}(-0.2, 0.2)$\\
& Joint position
  & $\mathcal{U}(-0.01, 0.01)$\\
& Joint velocity
  & $\mathcal{U}(-0.5, 0.5)$ \\

& \multicolumn{2}{@{}l}{\textit{Rope Morphology}} \\

& Segment position
  & $\mathcal{U}(-0.1, 0.1)$\\

& \multicolumn{2}{@{}l}{\textit{Player}} \\
& OBB position
  & $\mathcal{U}(-0.1, 0.1)$\\
& 6D OBB orientation
  & $\mathcal{U}(-0.05, 0.05)$\\
& OBB size
  & $\mathcal{U}(-0.1, 0.1)$ \\

\bottomrule
\end{tabular}
}
\label{tab:sim2real}
\end{table}

Table \ref{tab:sim2real} presents the dynamics randomization techniques and observation noises used in \MethodName. For each humanoid robot, we randomize its physical materials, body mass and center of mass in torso link and apply random push on base link. For discretely simulated rope, we randomize its length, density and internal drive properties. The observation terms that require on-board sensing or external MoCap streaming are injected with corresponding noises.

\section{Training Details}

Algorithm \ref{alg:training_procedure} displays the whole training procedure of \MethodName in different stages, where Line 1-9 shows the interleaved optimization on $\pi^{\text{jump}}$ and $\phi$ in Section \ref{sec:diverse_jump}, Line 11-16 presents the multi agent policy optimization on $\pi^{\text{manip}}$ in Section \ref{sec:low_level} and Line 18-24 corresponds to the high-level coordination in Section \ref{sec:high_level}. The training hyperparameters are summarized in Table \ref{tab:ppo_hyperparameters_low_level} \ref{tab:ppo_hyperparameters_high_level} \ref{tab:ppo_hyperparameters_player}.

\begin{table}[htb]
\centering
\caption{Training hyperparameters of low-level rope manipulation policy.}
\label{tab:ppo_hyperparameters_low_level}
\footnotesize
\setlength{\tabcolsep}{4pt}
\renewcommand{\arraystretch}{1.12}
\begin{tabular}{@{} ll @{\hspace{30pt}} l @{}}
\toprule
\textbf{Category} & \textbf{Hyperparameter} & \textbf{Value}\\
\midrule

\multirow{4}{2.8cm}{\textbf{Architecture}}
& Policy MLP hidden dimensions & $[512, 256, 128]$ \\
& Critic MLP hidden dimensions & $[512, 256, 128]$ \\
& Activation function & ELU \\
& Policy distribution & Gaussian \\
& Policy std range & $(0.1, 2.0)$ \\
& Observation history ($H$) & 5 \\
& Observable points on rope ($m$) & 8 \\ [2pt]
\midrule

\multirow{11}{2.8cm}{\textbf{Training}}
& Steps per environment
  & $25$\\
& Learning rate ($\eta$)
  & $3 \times 10^{-4}$\\
& Max gradient norm ($\|\mathbf{g}\|_{\text{clip}}$)
  & $1.0$\\
& Clip parameter
  & $0.2$\\
& Entropy coefficient
  & $0.01$\\
& Value loss coefficient
  & $1.0$\\
& Discount factor ($\gamma$)
  & $0.99$\\
& GAE $\lambda$
  & $0.95$\\
& Desired KL
  & $0.01$\\
& Learning epochs
  & $5$\\
& Mini-batches
  & $4$\\

\bottomrule
\end{tabular}
\end{table}

\begin{table}[htb]
\centering
\caption{Training hyperparameters of high-level scheduling policy.}
\label{tab:ppo_hyperparameters_high_level}
\footnotesize
\setlength{\tabcolsep}{4pt}
\renewcommand{\arraystretch}{1.12}
\begin{tabular}{@{} ll @{\hspace{30pt}} l @{}}
\toprule
\textbf{Category} & \textbf{Hyperparameter} & \textbf{Value}\\
\midrule

\multirow{5}{2.8cm}{\textbf{Architecture}}
& Policy MLP hidden dimensions & $[256, 128, 64]$ \\
& Critic MLP hidden dimensions & $[256, 128, 64]$ \\
& Activation function & ELU \\
& Policy distribution & Beta \\
& Concentration range & $[0.1, 25.0]$ \\
& Observation history ($H$) & 5 \\
\midrule

\multirow{11}{2.8cm}{\textbf{Training}}
& Steps per environment
  & $25$\\
& Learning rate ($\eta$)
  & $3 \times 10^{-4}$\\
& Max gradient norm ($\|\mathbf{g}\|_{\text{clip}}$)
  & $1.0$\\
& Clip parameter
  & $0.2$\\
& Entropy coefficient
  & $0.01$\\
& Value loss coefficient
  & $1.0$\\
& Discount factor ($\gamma$)
  & $0.99$\\
& GAE $\lambda$
  & $0.95$\\
& Desired KL
  & $0.01$\\
& Learning epochs
  & $5$\\
& Mini-batches
  & $4$\\
  
\bottomrule
\end{tabular}
\end{table}

\begin{table}[htb]
\centering
\caption{Training hyperparameters of diverse player jumping policy.}
\label{tab:ppo_hyperparameters_player}
\footnotesize
\setlength{\tabcolsep}{4pt}
\renewcommand{\arraystretch}{1.12}
\begin{tabular}{@{} ll @{\hspace{30pt}} l @{}}
\toprule
\textbf{Category} & \textbf{Hyperparameter} & \textbf{Value}\\
\midrule

\multirow{7}{2.8cm}{\textbf{Architecture}}
& Policy MLP hidden dimensions & $[512, 256, 128]$ \\
& Critic MLP hidden dimensions & $[512, 256, 128]$ \\
& Metric MLP hidden dimensions & $[256, 128, 64]$ \\
& Activation function & ELU \\
& Latent command dimension ($d_{\text{latent}}$) & $4$\\
& Policy distribution & Gaussian \\
& Standard derivation range & $(0.1, 2.0)$ \\
& Observation history ($H$) & 10 \\
\midrule

\multirow{13}{2.8cm}{\textbf{Training}}
& Steps per environment
  & $25$\\
& Learning rate ($\eta$)
  & $3 \times 10^{-4}$\\
& Max gradient norm ($\|\mathbf{g}\|_{\text{clip}}$)
  & $1.0$\\
& Clip parameter
  & $0.2$\\
& Entropy coefficient
  & $0.01$\\
& Value loss coefficient
  & $1.0$\\
& Discount factor ($\gamma$)
  & $0.99$\\
& GAE $\lambda$
  & $0.95$\\
& Desired KL
  & $0.01$\\
& Learning epochs
  & $5$\\
& Mini-batches
  & $4$\\
& Task reward threshold for diversity intrinsic
  & $0.3$\\
& Diversity intrinsic weight ($\beta$)
  & $1.0$\\

\bottomrule
\end{tabular}
\end{table}

\section{Reward Design}

\begin{table}[htb]
\caption{Reward terms for different policy training in \MethodName.}
\centering

\footnotesize
\setlength{\tabcolsep}{4pt}
\renewcommand{\arraystretch}{1.25}

\resizebox{\textwidth}{!}{
\begin{tabular}{@{} ll @{\hspace{18pt}} ll @{}}
\toprule
 & \textbf{Term} & \textbf{Expression} & \textbf{Weight} \\
\midrule

\multirow{16}{2.8cm}{\textbf{Low-level}\\ \textbf{Rope} \\ \textbf{Manipulation}}
& \multicolumn{3}{@{}l}{\textit{Task Objectives}} \\
& Track linear velocity
  & $\exp\!\left(-8.0\left\|\mathbf{v}_{xy}^{\text{c}} - \left[ \hat{v}_{x}^{\text{lin}}, \hat{v}_{y}^{\text{lin}} \right] \right\|_2^2\right)$
  & 2.0 \\
& Track angular velocity
  & $\exp\!\left(-8.0\left\|\omega_{z}^{\text{r}} - \hat{v}_{z}^{\text{ang}}\right\|_2^2\right)$
  & 2.0 \\
& Track width
  & $\exp\!\left(-20.0\left|\left\|\mathbf{p}^{\text{end}}_{xy}-\mathbf{p}^{\text{start}}_{xy}\right\|_2-\hat{w}\right|\right)$
  & 1.0 \\
& Track rotation
  & $\exp\!\left(-0.04\left\|\bar{\boldsymbol{\omega}} - \hat{\boldsymbol{\omega}}\right\|_2^2\right)$
  & 4.0 \\

& \multicolumn{3}{@{}l}{\textit{Regularization}} \\
& Face to other
  & $\left\| \Delta\theta_{\text{facing}} \cdot \boldsymbol{1}\left( \Delta\theta_{\text{facing}} < \frac{\pi}{12} \right) \right\|_2^2$
  & $-1.0$ \\
& Action rate
  & $\left\|a_t - a_{t-1}\right\|_2^2$
  & $-0.05$ \\
& Flat orientation
  & $\left\| \mathbf{g}_{xy}^{b} \right\|_2^2$
  & $-5.0$ \\
& Angular velocity in $xy$
  & $\left\|\boldsymbol{\omega}_{xy}^{b}\right\|_2^2$
  & $-0.05$ \\
& Joint position limits
  & $\sum_i\!\Bigl[\mathbf{1}(\mathbf{q}_{i} \ge \mathbf{q}_{i}^{\text{upper}}) + \mathbf{1}(\mathbf{q}_{i} \le \mathbf{q}_{i}^{\text{lower}})\Bigr]$
  & $-10.0$ \\
& Joint deviation (upper body)
  & $\sum_{i\in\mathcal{U}}\left|\mathbf{q}_{i} - \mathbf{q}_{i}^{\text{default}}\right|$
  & $-0.05$ \\
& Joint deviation (waist)
  & $\sum_{i\in\mathcal{W}}\left|\mathbf{q}_{i} - \mathbf{q}_{i}^{\text{default}}\right|$
  & $-0.1$ \\
& Joint deviation (lower body)
  & $\sum_{i\in\mathcal{L}}\left|\mathbf{q}_{i} - \mathbf{q}_{i}^{\text{default}}\right|$
  & $-0.1$ \\
& Slide with contact
  & $\sum_k \left\|\mathbf{v}_{xy}^{\text{foot}_k,t}\right\|_2\mathbf{1}(\text{foot}_k\text{ contacts})$
  & $-0.4$ \\
& Undesired contacts
  & $\mathbf{1}(\text{non-foot bodies contact})$
  & $-1.0$ \\

\midrule

\multirow{7}{2.8cm}{\textbf{High-level Scheduling}}
& \multicolumn{3}{@{}l}{\textit{Task Objectives}} \\
& Track frequency
  & $\exp\!\left(-1.5\left(\left\| \bar{\boldsymbol{\omega}} \right\| / 2 \pi - \nu \right)^2\right)$
  & 1.0 \\
& Track phase
  & $\exp\!\left(-1.0\left\| \bar{\theta}^{\text{rope}} - \hat{\theta}^{\text{rope}} \right\|_2^2\right)$
  & 1.0 \\
& Track player
  & $\exp\!\left(-4.0\left\|\mathbf{p}^{\text{c}}_{xy}-\mathbf{p}^{\text{player}}_{xy}\right\|_2^2\right)$
  & 1.0 \\
& Overlap
  & $\mathbf{1}\left(\exists k,\ \mathbf{p}^{\text{rope}}_{t}[k] \in \text{OBB}_{\text{player}}\right)$
  & $-1.0$ \\
& \multicolumn{3}{@{}l}{\textit{Regularization}} \\
& Command rate
  & $\left\|\mathbf{\mathbf{u}}_{t} - \mathbf{u}_{t - 1} \right\|_2^2, \mathcal{G}_{t}^{\text{manip}} = \operatorname{affine}(\mathbf{u}_{t})$
  & $-0.2$ \\

\midrule

\multirow{17}{2.8cm}{\textbf{Player}}
& \multicolumn{3}{@{}l}{\textit{Task Objectives}} \\
& Align rhythm
  & ($2 \cdot \mathbf{1}(\text{contact} = c) - 1) (0.8 \cdot \mathbf{1}(c = 1) + 0.2 \cdot \mathbf{1}(c = 0))$
  & 8.0 \\

& \multicolumn{3}{@{}l}{\textit{Regularization}} \\
& Feet distance
  & $d_{\text{feet}} - \operatorname{clip}(d_{\text{feet}}, 2.5, 4)$
  & $-2.0$ \\
& Reduce linear velocity
  & $\exp\!\left(-1.5\left\|\mathbf{v}^{\text{base}}_{xy}\right\|_2^2\right)$
  & 1.0 \\
& Reduce angular velocity
  & $\exp\!\left(-2.0\left\|\omega^{\text{base}}_{z}\right\|_2^2\right)$
  & 1.0 \\
& Action rate
  & $\left\|\mathbf{a}_t-\mathbf{a}_{t-1}\right\|_2^2$
  & $-0.05$ \\
& Angular velocity in $xy$
  & $\left\|\boldsymbol{\omega}^{xy}_{t}\right\|_2^2$
  & $-0.05$ \\
& Joint position limits
  & $\sum_i\!\Bigl[\mathbf{1}(\mathbf{q}_{i} \ge \mathbf{q}_{i}^{\text{upper}}) + \mathbf{1}(\mathbf{q}_{i} \le \mathbf{q}_{i}^{\text{lower}})\Bigr]$
  & $-10.0$ \\
& Joint deviation (upper body)
  & $\sum_{i\in\mathcal{U}}\left|\mathbf{q}_{i} - \mathbf{q}_{i}^{\text{default}}\right|$
  & $-0.1$ \\
& Joint deviation (waist)
  & $\sum_{i\in\mathcal{W}}\left|\mathbf{q}_{i} - \mathbf{q}_{i}^{\text{default}}\right|$
  & $-0.8$ \\
& Joint deviation (lower body)
  & $\sum_{i\in\mathcal{L}}\left|\mathbf{q}_{i} - \mathbf{q}_{i}^{\text{default}}\right|$
  & $-0.5$ \\
& Joint mirror (upper body)
  & $\left\|\mathbf{q}^{\text{L}}_{\text{upper}}-\mathbf{q}^{\text{R}}_{\text{upper}}\right\|_2^2$
  & $-0.2$ \\
& Joint mirror (lower body)
  & $\left\|\mathbf{q}^{\text{L}}_{\text{lower}}-\mathbf{q}^{\text{R}}_{\text{lower}}\right\|_2^2$
  & $-0.2$ \\
& Feet contact force
  & $\left| \operatorname{clip} \left( \frac{\mathbf{F}_{z}^{\text{feet}}}{m_{\text{body}}} - 2.5 g, 0, \infty \right) \right|_2$
  & $-0.1$ \\
& Slide with contact
  & $\sum_k \left\|\mathbf{v}_{xy}^{\text{foot}_k,t}\right\|_2\mathbf{1}(\text{foot}_k\text{ contacts})$
  & $-0.5$ \\
& Undesired contacts
  & $\mathbf{1}(\text{non-foot bodies contact})$
  & $-1.0$ \\

\bottomrule
\end{tabular}
}
\label{tab:reward_terms}
\end{table}

Table \ref{tab:reward_terms} lists the detailed reward terms designed for specific policy in \MethodName. Specifically, for low-level rope manipulation, four main reward terms are used to guide policy $\pi^{\text{manip}}$ to follow the command $\mathcal{G}^{\text{manip}}$. Additional regularization terms penalize wrong facing direction, unstable control and overall poses. For high-level scheduling, besides the phase tracking reward mentioned in Section \ref{sec:high_level}, we also introduce an auxiliary shaping reward for rope to rotate under the clock frequency for faster convergence. The player tracking reward and overlap penalty incentivize $\pi^{\text{sched}}$ to adapt to the slight horizontal drift of player and reduce the rope-player collision. For player jumping policy training, a single rhythm alignment reward is enough to derive the periodic jumping behavior as described in Section \ref{sec:high_level}. To prevent weird motion style when adding diversity intrinsic, the rest regularization terms further constrains the feet distance, joint symmetry, feet contact force, etc.

\end{document}